\documentclass[11pt,letter]{article}
\usepackage{times,latexsym}
\usepackage{url}
\usepackage[T1]{fontenc}

\usepackage[acceptedWithA]{tacl2021v1}

\usepackage{xspace,mfirstuc,tabulary}

\usepackage{graphicx}

\usepackage{algorithmic,algorithm}
\renewcommand{\algorithmiccomment}[1]{\bgroup\hfill$\triangleright$~#1\egroup}
\usepackage[linguistics]{forest} 
\usepackage{synttree}
\usepackage{subcaption}
\usepackage{tikz-dependency}
\usepackage[cjk]{kotex}

\usepackage{amsmath,amssymb}

\usepackage{arydshln}
\usepackage{booktabs}

\usepackage{amsthm}

\newif\iftaclinstructions
\taclinstructionsfalse 
\iftaclinstructions
\renewcommand{\confidential}{}
\renewcommand{\anonsubtext}{(No author info supplied here, for consistency with
TACL-submission anonymization requirements)}
\newcommand{\instr}
\fi

\iftaclpubformat 

\else

\fi

\title{Learning Constituent Headedness}

\author{
Zeyao Qi$^{1\dagger}$~~~~ Yige Chen$^{1\dagger}$~~~~ KyungTae Lim$^{2}$~~~~ Haihua Pan$^{1*}$~~~~ Jungyeul Park$^{2}$\thanks{~~Corresponding authors. $\dagger$Equally contributed authors.}\\
$^{1}$The Chinese University of Hong Kong\\
$^{2}$Korea Advanced Institute of Science \& Technology\\
\url{https://ling.cuhk.edu.hk}~~~~ \url{https://ct.kaist.ac.kr}
}
\date{}

\begin{document}
\maketitle

\begin{abstract}
Headedness is widely used as an organizing device in syntactic analysis, yet constituency treebanks rarely encode it explicitly and most processing pipelines recover it procedurally via percolation rules.
We treat this notion of constituent headedness as an explicit representational layer and learn it as a supervised prediction task over aligned constituency and dependency annotations, inducing supervision by defining each constituent head as the dependency span head.
On aligned English and Chinese data, the resulting models achieve near-ceiling intrinsic accuracy and substantially outperform Collins-style rule-based percolation.
Predicted heads yield comparable parsing accuracy under head-driven binarization, consistent with the induced binary training targets being largely equivalent across head choices, while increasing the fidelity of deterministic constituency-to-dependency conversion and transferring across resources and languages under simple label-mapping interfaces.
\end{abstract}

\section{Introduction}

Head information is routinely used as an interface signal in computational syntax.
It drives head-driven binarization \citep{collins-1999-head,charniak-1997-statistical}, supports tree normalization \citep{bikel:2004:CL}, and mediates conversion between constituency and dependency representations \citep{de-marneffe-etal-2006-generating}.
Despite this central role, most constituency treebanks do not encode heads explicitly, and practical pipelines recover them procedurally through hand-designed percolation rules.
As a consequence, head selection is rarely evaluated as an object in its own right, and errors are difficult to analyze independently of the downstream transformation that happens to use the heads.

In this paper, we treat headedness as an explicit \emph{interface layer} between constituency and dependency representations.
Concretely, headedness denotes the choice of a head child for each nonterminal that makes aligned constituency and dependency representations invertible under an alignment criterion.
This definition does not aim to reconstruct theoretical headedness as studied in syntax \citep{chomsky-1970-remarks,jackendoff-1977-x-syntax,carnie-2002-syntax}; it targets the practical head decisions that allow deterministic structural operations to be carried out consistently across annotation schemes.

Existing systems typically introduce constituent heads procedurally.
Many parsers rely on Collins-style percolation tables or related rule inventories, and these rules are often embedded as auxiliary mechanisms for lexicalization, feature extraction, or head-driven actions \citep{sagae-lavie-2005-classifier,zhu-etal-2013-fast,watanabe-sumita-2015-transition}.
Even when head selection is integrated into a model component, its quality is usually assessed indirectly through parsing accuracy rather than by direct comparison to an explicit head standard \citep{hou-li-2025-dynamic}.
{This procedural status can have practical drawbacks: percolation tables are typically hand-tuned to a particular treebank's nonterminal inventory and rule priorities, which can make cross-resource comparison opaque, and diagnosing conversion or binarization errors requires reasoning about both the rules and the downstream transformation simultaneously.}

{To address the aforementioned issue, we} make headedness directly observable, learnable, and evaluable.
We exploit treebanks in which constituency and dependency annotations are available for the same sentences.
Dependency trees provide head–dependent relations among terminals \citep{hajic-etal-2003-prague}, while constituency trees provide hierarchical grouping \citep{marcus-etal-1993-building}.
Aligning the two, we induce gold constituent heads by defining the head of each constituent as its dependency span head, yielding supervision that does not depend on hand-crafted percolation conventions.

Using these induced annotations, we show that constituent headedness can be learned with near-ceiling intrinsic accuracy on aligned English and Chinese data, substantially outperforming rule-based percolation.
We then test whether making headedness explicit matters for downstream use cases that rely on head information.
Under head-driven binarization with a fixed parsing architecture, predicted heads yield parsing accuracy comparable to rule-based heads, consistent with the induced binary training targets being largely equivalent up to local spine choices.
For deterministic constituency-to-dependency conversion, predicted heads markedly increase recovery fidelity relative to rule-based heads.
Finally, we evaluate transfer across resources and across languages under simple label-mapping interfaces, characterizing when this headedness layer generalizes and where resource-specific conventions limit portability.

Overall, the paper repositions constituent headedness from a parser-internal heuristic to an explicit, empirically testable interface component.
By defining headedness through aligned supervision and evaluating it both intrinsically and through controlled downstream transformations, we provide a practical alternative to rule-based percolation for resource-robust binarization and cross-formalism conversion.

\section{Background and related work}

This section provides the representational background needed to state the headedness task precisely and situates it within prior work on head specification in parsing and treebank processing.
The emphasis is on headedness as a property that is frequently required by downstream procedures, while remaining largely implicit in standard constituency representations.

\subsection{Headedness as missing structure in constituency trees}

Standard constituency trees encode hierarchical grouping and category labels but do not explicitly mark which child of a constituent functions as its head.
As a result, headedness is not part of the observable tree representation: multiple head assignments may be compatible with the same labeled bracket structure, even when the local configuration and yield are fixed.
This representational underspecification becomes consequential {when} a procedure requires a distinguished child to mediate structural operations over the tree, including binarization, lexicalization, and systematic mappings between constituency and dependency representations.

\subsection{Head percolation and head driven binarization}

Head information has long been operationally important in bottom-up parsing, particularly in CKY-style dynamic programming \citep{cocke:1969,kasami:1966,younger:1967}, where binarization is used to control rule expansion and parsing complexity.
In widely used treebank pipelines, heads are typically supplied by hand-crafted head percolation rules, most prominently those introduced by Collins \citep{collins:1996:ACL,collins-1999-head}.
Such rule tables define deterministic mappings from a parent category and an ordered sequence of children to a single head child.
They have been adopted as a practical interface for head-driven binarization and lexicalized parsing, and they remain a standard mechanism for injecting head information into otherwise headless phrase structure trees.

Although effective within a fixed resource, percolation schemes are tightly coupled to label inventories and rule ordering, and they encode annotation conventions rather than a tree-internal notion of headhood.
Consequently, head choice is often treated as a procedural artifact of a particular pipeline rather than as an independently evaluable layer of representation.

\subsection{Conversion between constituency and dependency}

{A second major use of head information is deterministic conversion between constituency and dependency representations.}
Because dependency trees encode explicit head--dependent relations among terminals, conversion procedures typically require a head choice at each constituent in order to project dependency relations from phrase structure.
This perspective has motivated rule-based conversion pipelines in which constituent heads are recovered via percolation and then used to construct dependency arcs, as in early conversion work on Penn-style treebanks \citep{xia-palmer-2001-converting,levy-manning-2004-deep}.
Conversely, dependency representations have also been derived from constituency annotations in widely used toolchains, including Stanford-style conversions \citep{de-marneffe-etal-2006-generating}.
Across these approaches, head selection is a critical intermediate decision: errors in head assignment propagate to attachment choices and can affect the reversibility and fidelity of conversion.

\subsection{Learning heads inside parsing architectures}

Beyond hand-designed percolation tables, a related line of work treats headedness as part of the parsing mechanism rather than as a separately learned representational layer.
In particular, several discriminative and transition-based constituency parsers rely on predefined head rules to support lexicalization and feature extraction, or to define head-driven parsing actions; head selection is therefore used during parsing but is not itself learned from head-annotated supervision \citep{sagae-lavie-2005-classifier,zhu-etal-2013-fast,watanabe-sumita-2015-transition}.
A dynamic head selection mechanism can also be integrated into a lexicalized neural parser, where head choices are made online to enable lexicalized rule expansion \citep{hou-li-2025-dynamic}.
Across these architectures, the contribution of headedness is typically assessed indirectly through end-to-end parsing accuracy, rather than through direct evaluation of predicted head assignments against an explicit gold standard for constituent heads.

\subsection{Summary and contrast}

Prior work therefore uses heads primarily as an enabling device: percolation rules supply heads to support binarization and conversion, and learning-based methods select heads as part of a parser-internal decision process.
In contrast, {our} work treats constituent headedness as an explicit representational layer over gold constituency trees and evaluates it directly.
By inducing gold head annotations from constituency--dependency alignment and formulating head identification as an independent prediction task, we separate headedness from parser-specific mechanisms and assess its role as a reusable interface for head-driven binarization and cross-formalism conversion.

\section{Problem definition}

This section formalizes constituent headedness as an explicit property of constituency representations and defines the task of head identification independently of any particular parsing architecture.

\subsection{Constituent headedness as a representational property}

We assume a rooted constituency tree $T=(V,E)$ whose terminals form a token sequence $w_1,\ldots,w_n$.
Let $V_{\mathrm{NT}} \subseteq V$ be the set of non-terminal nodes, and let each $v\in V_{\mathrm{NT}}$ dominate an ordered sequence of immediate children
$\mathit{ch}(v)=\langle v_1,\ldots,v_k\rangle$.
Let $\mathit{yield}(v)\subseteq \{1,\ldots,n\}$ denote the set of terminal indices dominated by $v$.

Constituent headedness is defined as a function
\[
h: V_{\mathrm{NT}} \rightarrow V
\]
that assigns exactly one head child $h(v)\in \mathit{ch}(v)$ to each non-terminal node $v\in V_{\mathrm{NT}}$.
The head child is intended to represent the element from which the constituent projects its core syntactic properties.
Under this formulation, headedness is a structural property defined over the constituency tree itself and can be specified independently of any parsing procedure.

A constituency tree augmented with head annotations therefore constitutes a strictly richer representation than an unlabeled tree, while preserving the original hierarchical organization.
This explicit representation makes headedness directly observable and amenable to independent evaluation.

\subsection{Head prediction task}

Given a gold constituency tree $T$ without head annotations, the task is to predict the head function $h$ for all non-terminal nodes.
For a constituent with $k$ immediate children, head prediction corresponds to a constrained $k$-way classification problem in which exactly one child must be selected:
\[
\sum_{i=1}^{k} \mathbb{I}[h(v)=v_i] = 1 
\]
No constituent may have zero or multiple heads.
The task is defined over gold constituency trees rather than parser outputs, ensuring that head prediction is evaluated independently of parsing accuracy.

\subsection{Deriving gold headedness from aligned dependency trees}

The formulation above treats headedness as a representational property, but it does not specify how gold head annotations are obtained.
To make headedness observable, we consider sentences for which a constituency tree $T$ and a dependency tree $D$ are jointly available and aligned at the terminal level.

Let $D$ be a rooted dependency tree over $\{0,1,\ldots,n\}$ with head function
$g:\{1,\ldots,n\}\rightarrow \{0,1,\ldots,n\}$ assigning a unique governor to each token
(with $0$ the artificial root).
For each non-terminal node $v\in V_{\mathrm{NT}}$, consider the set of span-head candidates
\[
H(v) \;=\; \{\, i \in \mathit{yield}(v) \;:\; g(i) \notin \mathit{yield}(v) \,\}.
\]
When $|H(v)|=1$, we define the dependency head token of $v$ as
\[
\mathit{hd}(v) \;=\; \text{the unique } i \in H(v).
\]
Intuitively, $\mathit{hd}(v)$ is the token in the constituent yield whose dependency parent lies outside the constituent, i.e., the attachment point of the span in the dependency structure.
We then define the head child of $v$ as the unique child that dominates this token:
\[
\begin{aligned}
h(v) \;=\; \text{the unique } v_j \in \mathit{ch}(v)\ \text{such that}\\
\mathit{hd}(v)\in \mathit{yield}(v_j).
\end{aligned}
\]
This yields a single head child whenever $T$ and $D$ are aligned and $|H(v)|=1$ holds, and it is independent of phrase labels, linear precedence, and language-specific percolation rules.
Sentences for which the uniqueness condition fails for any $v\in V_{\mathrm{NT}}$ are excluded from the aligned subset used for supervision.

\section{Experiments on aligned treebanks}

This section evaluates constituent headedness as an explicit representational layer over constituency trees.
The goal is not to introduce new parsing architectures or objectives, but to isolate the contribution of head annotations under controlled interfaces in which all non-head factors are held fixed.

We use three complementary evaluations that target distinct consequences of making headedness explicit.
Experiment~1 is an intrinsic test of the head function $h$, measuring head identification accuracy on gold constituency trees.
Experiment~2 evaluates headedness as a structural control signal by comparing head-driven binarization under a fixed parsing architecture, varying only the source of head annotations.
Experiment~3 evaluates headedness as a conversion interface by measuring fidelity under deterministic constituency--dependency mappings, so that differences can be attributed directly to head assignment rather than model variability.

All experiments are conducted on aligned subsets of the Penn Treebank for English and a Chinese constituency treebank with corresponding dependency annotations \citep{marcus-etal-1993-building,xue-etal-2005-ctb}. For English, we use the standard Penn Treebank split: Sections~02–21 for training, Section~22 for development, and Section~23 for testing. For Chinese, we follow the predefined {train/dev/test} splits used for constituency parsing in prior work.\footnote{\url{https://github.com/nikitakit/self-attentive-parser/tree/master/data/ctb_5.1}} 
{After normalization, constituency and dependency structures are exactly aligned for every sentence except one Chinese training instance with an apparent constituency-side annotation error, which we exclude.} 
We otherwise retain only sentences whose constituency and dependency representations align, so that gold head annotations derived from dependency structure are well-defined and unambiguous. This filtering prioritizes reliability of supervision over corpus coverage and decouples headedness evaluation from tokenization mismatches.

\subsection{Experiment~1: intrinsic evaluation of head prediction} \label{experiment-1}

Experiment~1 evaluates constituent headedness as an intrinsic prediction problem, independent of any downstream parsing objective.
Given a gold constituency tree, the model predicts the head child of each non-terminal node, and we compute accuracy as the proportion of constituents whose predicted head child matches the alignment-derived gold head child.
We report overall accuracy and breakdowns by major phrase category.

Gold head children are induced from sentence-level alignments between gold constituency and dependency trees.
For each non-terminal node $v$, let $\mathit{yield}(v)$ denote the set of terminal positions dominated by $v$.
We define the dependency span head $\mathit{hd}(v)$ as the unique token in $\mathit{yield}(v)$ whose dependency governor lies outside $\mathit{yield}(v)$ (counting the artificial root as outside).
We restrict evaluation to nodes for which $\mathit{hd}(v)$ is well-defined and unique under the alignment.
The gold head child for $v$ is then the unique child $v_j \in \mathit{ch}(v)$ such that $\mathit{hd}(v) \in \mathit{yield}(v_j)$.
Each training instance consists of a parent node and its ordered children, paired with the $1$-based index of the gold head child.

We use an unlexicalized input representation: the encoder receives only the parent category/tag and the ordered child categories/tags, with no sentence words or other lexical features.
To ensure feature consistency across treebanks, we normalize category and tag strings by removing redundant functional or morphological suffixes.

Head prediction is modeled as a constrained $k$-way classification problem over the immediate children of each constituent, where $k = |\mathit{ch}(v)|$ varies by node.
Our model uses a Transformer encoder \citep{devlin-etal-2019-bert} followed by a shallow multilayer perceptron that predicts the head-child index.
The model is evaluated on gold constituency structure and does not perform parsing.
As a rule-based baseline, we apply standard head-percolation procedures and evaluate them against the same alignment-derived gold heads, using the Collins-style rules in {Stanford CoreNLP} for English and Chinese.

Across both English and Chinese, learned headedness achieves near-ceiling accuracy and substantially outperforms rule-based head selection (Table~\ref{head-accuracy}).
Full implementation details and hyperparameters are provided in Appendix~\ref{experiment-1-detail}.

\begin{table}
\centering
\footnotesize
\resizebox{.48\textwidth}{!}{
\begin{tabular}{r cc} \toprule
& Rule-based headedness & Learned headedness \\ \midrule
English  &67.89  &98.54  \\
Chinese  &92.42  &100.00  \\ 
\bottomrule
\end{tabular}
}
\caption{Head prediction accuracy for constituent headedness using rule-based and learned approaches.}
\label{head-accuracy}
\end{table}

\subsection{Experiment~2: head-driven binarization and parsing stability} \label{experiment-2}

Experiment~2 tests whether constituent head choice affects constituency parsing when heads are used only to guide head-driven binarization, with the parsing architecture held fixed.
The target property is \emph{stability}: if different head conventions primarily rotate the internal binary spines introduced for training, then parsers trained under different heads should achieve comparable accuracy once predictions are deterministically debinarized.

Let $\mathcal{T}$ denote the original gold constituency trees.
We first apply {a deterministic punctuation-aware normalization procedure} to obtain $\mathcal{T}^{\mathrm{norm}}$.
We then apply the same head-driven binarization operator under two head sources, yielding two binarized training sets:

{\footnotesize
\[
\mathcal{T}^{\mathrm{bin}}_{\mathrm{rule}}
\;=\;
\mathsf{Bin}\!\left(\mathcal{T}^{\mathrm{norm}}; H_{\mathrm{rule}}\right), \mathcal{T}^{\mathrm{bin}}_{\mathrm{learn}}
\;=\;
\mathsf{Bin}\!\left(\mathcal{T}^{\mathrm{norm}}; H_{\mathrm{learn}}\right)
\]}
where $H_{\mathrm{rule}}$ is derived from Collins-style percolation rules and $H_{\mathrm{learn}}$ is predicted by our headedness model.
In both conditions, the normalization procedure, binarization operator $\mathsf{Bin}$, parsing architecture, data splits, and hyperparameters are identical.

For each condition, we train a constituency parser on $\mathcal{T}^{\mathrm{bin}}_{\mathrm{rule}}$ or $\mathcal{T}^{\mathrm{bin}}_{\mathrm{learn}}$.
At test time, the parser predicts a binarized tree $\mathcal{P}^{\mathrm{bin}}$, which is deterministically debinarized to obtain $\mathcal{P}$.
We evaluate $\mathcal{P}$ against the original gold tree $\mathcal{T}$ using labeled bracket $F_1$, computed with \textsc{jp-evalb} \citep{jo-etal-2024-novel}, which explicitly includes punctuation during constituent evaluation.

Table~\ref{parsing-accuracy} shows that replacing rule-based heads with learned heads yields only small differences in labeled bracket $F_1$ on both English and Chinese.
Under our significance test, these differences are not statistically {significant (English: $p=0.636$; Chinese: $p=0.497$)}, and the corresponding effect sizes are small (English: $\Delta F_1=0.08$; Chinese: $\Delta F_1=0.36$).
This supports the stability interpretation: head choice changes the latent binary training targets but has limited impact on the debinarized evaluation target.

\begin{table}
\centering
\footnotesize
\resizebox{.48\textwidth}{!}{
\begin{tabular}{r cc} \toprule
& Rule-based headedness & Learned headedness \\ \midrule
English  &96.24  &96.32  \\
Chinese  &93.84  &94.20  \\
\bottomrule
\end{tabular}
}
\caption{Parsing accuracy (labeled bracket $F_1$) under head-driven binarization using rule-based and learned headedness, with the parsing architecture held fixed.}
\label{parsing-accuracy}
\end{table}

To relate this outcome to the training signal, we directly compare the two binarized training treebanks $\mathcal{T}^{\mathrm{bin}}_{\mathrm{rule}}$ and $\mathcal{T}^{\mathrm{bin}}_{\mathrm{learn}}$ over the same underlying normalized trees $\mathcal{T}^{\mathrm{norm}}$.
On the training portion, the induced bracket sets overlap heavily (English: Bracketing $F_1=95.31$; Chinese: Bracketing $F_1=97.63$), even though exact spine identity is much lower (English: $38.75\%$ complete match; Chinese: $68.71\%$ complete match).
This combination is diagnostic: the two head sources often select different local spine realizations, but these divergences largely reduce to internal rotations within the same original constituent and rarely survive deterministic debinarization.
Consequently, both conditions supply the parser with nearly the same bracket supervision distribution, and large downstream accuracy gaps are not expected.
The overall pattern is consistent with reports that varying head selection inside parsing pipelines typically yields only marginal changes in parsing accuracy \citep{hou-li-2025-dynamic}.
Full preprocessing, binarization, training, and the structural-equivalence computation are provided in Appendix~\ref{experiment-2-detail}.

\subsection{Experiment~3: constituency--dependency conversion fidelity}

Experiment~3 evaluates headedness as an explicit interface for deterministic mapping {from} constituency {to} dependency representations, independently of parsing.
Starting from gold constituency trees, we apply a fixed constituency-to-dependency conversion procedure parameterized only by constituent head choices.
Given a head function $h$, we derive a dependency tree by assigning each token its governor according to the induced head--dependent relations licensed by the constituency structure and the selected head child at each constituent.
The resulting dependency structures are evaluated intrinsically using unlabeled attachment score (UAS).


We compare two conditions that differ only in the head information used by the conversion procedures: rule-based heads derived from Collins-style percolation rules versus heads predicted by the learned headedness model.
Because {conversions} are deterministic and all non-head components of the pipeline are held fixed, differences in UAS can be attributed directly to the quality of head assignment rather than to model variability.

Across both English and Chinese, conversions based on learned heads yield substantially higher UAS in constituency-to-dependency conversion than conversions based on rule-derived heads (Table~\ref{conversion-accuracy}).
Full conversion details are provided in Appendix~\ref{experiment-3-detail}.

\begin{table}
\centering
\footnotesize
\resizebox{.48\textwidth}{!}{
\begin{tabular}{r cc} \toprule
& Rule-based headedness & Learned headedness \\ \midrule
English  & 51.39 & 94.99 \\
Chinese  & 77.73 & 92.22 \\
\bottomrule
\end{tabular}
}
\caption{Conversion fidelity under constituency--dependency mapping on aligned English and Chinese treebanks. UAS evaluates constituency-to-dependency conversion.}
\label{conversion-accuracy}
\end{table}

The results of this experiment show that explicit constituent headedness supports faithful structural mapping from constituency to dependency representations.
High UAS indicates that dependency structures derived from constituency trees preserve correct head--dependent relations.
The strong correspondence between the two metrics suggests that accurate head assignment provides sufficient information to mediate between formalisms under deterministic conversion procedures.
In contrast, rule-based headedness leads to systematic degradation in UAS, reflecting error propagation from incorrect head choices.
These findings indicate that headedness is not merely a procedural aid for conversion, but a structurally central representational layer that constrains both dependency relations and constituent organization.


\section{Extending headedness across resources}
\label{sec:transfer}

The experiments above establish that constituent headedness is learnable under aligned supervision within a fixed resource.
This section tests whether a learned head selector generalizes across annotation resources and across languages, without introducing language-specific head rules.

\subsection{Transfer setting}

Let $\mathcal{R}_s$ and $\mathcal{R}_t$ be a source and target treebank resource, each providing constituency trees over token sequences.\footnote{When the target resource also provides dependency trees aligned to the constituency terminals, we evaluate transferred headedness using the same dependency-grounded head criterion as in Section~\ref{experiment-1}.}
A head predictor trained on $\mathcal{R}_s$ defines a function that maps each local constituency configuration to a head-child index.
Cross-resource transfer is non-trivial because resources may differ in their inventories of phrase categories, part-of-speech tags, and annotation conventions.

We therefore model transfer as a deterministic interface between inventories.
Let $\Sigma^{\mathrm{ph}}_s$ and $\Sigma^{\mathrm{ph}}_t$ denote the source and target phrase-label sets, and let $\Sigma^{\mathrm{pos}}_s$ and $\Sigma^{\mathrm{pos}}_t$ denote the corresponding part-of-speech inventories.
We define mapping functions $\phi:\Sigma^{\mathrm{ph}}_t \rightarrow \Sigma^{\mathrm{ph}}_s$ and 
$\psi:\Sigma^{\mathrm{pos}}_t \rightarrow \Sigma^{\mathrm{pos}}_s$, which normalize target labels into the source label space.\footnote{The mappings $\phi$ and $\psi$ are defined by deterministic alignment rules (e.g., direct matches, coarse category collapsing, and a small number of hand-specified correspondences) and are fixed for all transfer experiments.}
Given a target constituent $v$ with parent label $\ell(v)$ and ordered children $\mathit{ch}(v)=\langle v_1,\ldots,v_k\rangle$ with associated child labels, the transferred predictor is applied to the normalized configuration obtained by applying $\phi$ and $\psi$ to the target labels.
The predicted head child is then interpreted back in the original target tree as an index in $\{1,\ldots,k\}$.

This interface makes the transfer protocol explicit: the learned parameters are held fixed, and only the deterministic label mapping varies across resource pairs.
No retraining is performed on target data unless stated otherwise.

\subsection{Transfer conditions}

We evaluate transfer in {two} settings.
First, we transfer a head predictor trained on the Penn Chinese Treebank to the Sinica Treebank \citep{huang-etal-2000-sinica}.
This condition tests cross-resource generalization within Chinese, where constituency structures are broadly comparable but {label inventories} and annotation conventions differ.

{Second, we independently transfer two head predictors trained on the Penn Treebank and the {Penn Chinese Treebank} to the French Treebank \citep{abeille-clement-toussenel:2003}. This condition tests cross-lingual generalization from English to French and from Chinese to French under a shared constituency formalism and comparable phrase-structural primitives.}

\subsection{Evaluation}

Transferred headedness is evaluated with the same head-finding criterion used in the aligned setting whenever target dependency annotations are available and aligned to constituency terminals.
In that case, for each constituent $v$, we compare the transferred head child against the gold head child induced by the dependency span-head definition in Section~\ref{experiment-1}.
This yields an intrinsic head selection accuracy on the target resource, directly comparable to Experiment~1.

\begin{table}
\centering
\footnotesize
{
\begin{tabular}{ cc} \toprule
 Transfer setting & Head accuracy \\ \midrule
 CTB $\to$ Sinica & 69.41 \\
 PTB $\to$ FTB & 54.95 \\
 CTB $\to$ FTB & 64.70 \\
\bottomrule
\end{tabular}
}
\caption{Cross-resource and cross-lingual transfer of constituent headedness. Accuracy is computed against dependency-induced gold heads when aligned dependency annotations are available on the target resource.}
\label{across-resources-results}
\end{table}


{Table~\ref{across-resources-results} shows that within-language cross-resource transfer from CTB to Sinica is reasonably strong, despite the fact that the two Chinese treebanks {differ in annotation conventions}. This suggests that a substantial portion of constituent headedness is recoverable and generalizes across annotation sources within the same language. Cross-lingually, CTB $\to$ FTB is also comparatively effective, and notably outperforms PTB $\to$ FTB, indicating that transfer success is driven less by ``same vs. different language'' than by the compatibility of headedness between the source and target resources.}

{A closer look at error patterns reveals that prepositional phrases (PPs) are the dominant bottleneck for PTB $\to$ FTB transfer: PP accuracy collapses to 0.09\%, whereas CTB $\to$ FTB achieves 89.81\% on PPs. This aligns with the head-position preferences in each resource: PTB PPs are overwhelmingly head-final (97.98\%), while FTB and CTB PPs are almost categorically head-initial (99.31\% and 97.23\%). {Concretely}, French PPs like [\textsubscript{PP} \textit{dans} [\textsubscript{NP} \textit{la maison}]] and Chinese PPs like [\textsubscript{PP} \textit{zài} [\textsubscript{NP} \textit{jiā-lǐ}]] select the adposition/coverb as the head child, whereas the PTB dependency-induced headedness for English [\textsubscript{PP} \textit{in} [\textsubscript{NP} \textit{the house}]] selects the nominal complement as head, reversing the directionality. Since PPs constitute roughly a quarter of the evaluated constituents, this single systematic discrepancy largely explains the degraded PTB $\to$ FTB performance.}
{More broadly, the results highlight a key generalizability point: failures under transfer can arise even when grammars are comparable, because headedness is partially an annotation contract between constituency and dependency schemes.}


\section{Discussion}

Our experiments treat constituent headedness as an explicit layer over a fixed constituency tree: for each non-terminal node, select the unique child that dominates the span's dependency attachment point.
A central consequence of this design is that the ``gold head'' we learn is dependency-grounded. It is therefore directly appropriate for the two pipeline interfaces we evaluate (head-driven binarization and deterministic constituency--dependency conversion), but it is also explicitly tied to the dependency scheme (and its head conventions) used in a given resource.

Because supervision is induced from dependency dominance, the model is not trying to recover an abstract, theory-internal head notion; it is learning a treebank-specific projection convention that makes constituency spans compatible with the resource's dependency analyses.
{This distinction matters for interpretation: Collins-style percolation provides a reasonable symbolic default, but it is less flexible than a learned head chooser when linguistically motivated conventions vary across resources. Under such shifts, disagreements with a Collins table primarily reflect the limited adaptivity of fixed rules, and the stronger performance of the learned model reflects its ability to fit the target resource without implying that one choice is linguistically wrong.}
Framing headedness as an explicit, supervised layer makes these conventions observable and comparable rather than buried inside a conversion script or rule table.

On the aligned subsets, the dependency-induced head child is determined by a structural criterion (unique outgoing governor from the span).
The fact that our model approaches ceiling indicates that, for these resources and under this criterion, headedness is largely recoverable from local syntactic information: the parent label and the ordered configuration of child labels/POS tags.
Compared to hand-written percolation tables, a learned model can fit fine-grained, resource-specific regularities over entire child sequences, and can capture interactions that do not factor into a simple priority list of child categories per parent.
In other words, the accuracy gain does not rely on lexical identity; it comes from learning how this treebank resolves frequent local configurations to match its dependency scheme.

In Experiment~2, improved heads yield modest but consistent gains in labeled bracketing \(F_1\) under controlled head-driven binarization.
This is a realistic effect size: modern constituency parsers can often learn good structure even when binarization is not perfectly aligned with the target representation, and binarization is only one component of the training signal.
In contrast, Experiment~3 shows very large UAS differences in deterministic constituency-to-dependency conversion.
This asymmetry is diagnostic. Conversion uses headedness as a hard interface. A wrong head at an internal node can invert attachment direction and propagate to multiple token-level arcs.
Thus, percolation choices that may be ``good enough'' for certain parsing pipelines can still be a poor match for a dependency standard, while heads learned to match the dependency-induced criterion naturally yield high-fidelity conversion.

Making headedness explicit provides a reusable phrase-to-token anchor that is often needed but usually reconstructed heuristically.
Even when downstream models do not manipulate trees directly, head annotations can support interfaces that require choosing a single token anchor for a span (e.g., aligning span-level predictions to token-level structures or defining consistent anchoring for head-driven span representations).
The empirical takeaway from Experiment~1 is that such anchors can be learned accurately under aligned supervision, offering a principled alternative to hand-written heuristics when a pipeline requires head-anchored operations.

\section{Conclusion}

{We argue that constituent headedness should be treated as an explicit representational layer over constituency trees, rather than as an unobserved procedural artifact supplied by hand-written percolation rules.
Using treebanks with aligned constituency and dependency annotations, we induce gold head children via the dependency span-head criterion and trained a supervised head predictor over gold constituency structure.}

{Empirically, learned headedness achieves near-ceiling intrinsic accuracy on aligned English and Chinese and substantially outperforms Collins-style percolation.
Using predicted heads as the sole varying signal in downstream pipelines, we showed that they (i) yield modest but consistent gains when used for head-driven binarization under a fixed constituency parsing architecture, and (ii) dramatically improve deterministic constituency-to-dependency conversion fidelity.
Finally, we found that headedness partially transfers across resources and languages under simple label mapping, with transfer success largely explained by the compatibility of head conventions.}

{These results support a practical conclusion for treebank processing and syntax toolchains: when headedness is needed as an interface for downstream tasks, it can be learned directly from aligned supervision, evaluated explicitly, and ported across datasets more transparently than rule tables.}
{Future work can further strengthen cross treebank and cross lingual generalization by explicitly modeling systematic convention differences such as head directionality and function word attachment, and by evaluating headedness as a transferable interface for downstream tasks beyond parsing and conversion.}


\appendix

\section{Experiment 1 details} \label{experiment-1-detail}

This section describes the construction of training instances and the classifier used in Experiment~1.

\paragraph{Instance construction}
We implement a head-induction procedure that maps each constituency node to a dependency-grounded span head and then to a head-child index.
For each non-terminal node, we compute its absolute token span $\mathit{yield}(v)$ and identify $\mathit{hd}(v)$ as the unique token in the span whose dependency governor lies outside the span.
We then map $\mathit{hd}(v)$ to the unique child in $\mathit{ch}(v)$ that dominates it, and record the corresponding 1-based ordinal position.
Each instance is represented by the parent category (or tag), the ordered sequence of child categories (or tags), and the gold head-child index.
All labels are normalized by removing redundant morphological or functional suffixes.

\paragraph{Classifier and input formatting}
We use \texttt{bert-base-uncased} as the encoder and a two-layer multilayer perceptron prediction head with a 256-dimensional hidden layer, ReLU activation, and dropout $p=0.2$.
Inputs are formatted as a pair of segments to allow the encoder to condition on the parent and the ordered children:
\[
X = \text{[CLS]} \; w \; \text{[SEP]} \; f_1, f_2, \ldots, f_m \; \text{[SEP]},
\]
where $w$ is the parent tag and $f_1,\ldots,f_m$ are the child tags.
Sequences are truncated to a maximum length of 64 tokens.
Training is performed for 10 epochs using AdamW with learning rate $2\times 10^{-5}$ and batch size 16 on an NVIDIA A100 GPU.
We select the final checkpoint by minimum development loss.

\section{Experiment 2 details} \label{experiment-2-detail}

This section describes the punctuation-aware normalization, head-driven binarization, and parser training setup used in Experiment~2.

\paragraph{Punctuation-aware normalization}
We introduce a preprocessing stage that integrates punctuation marks into the constituent structure prior to binarization.
A recursive traversal identifies punctuation tokens using predefined inventories of left and right delimiters.
When an opening delimiter (e.g., \texttt{(} or \texttt{<}) is encountered, the procedure searches for the corresponding closing delimiter according to a delimiter-pair map.
If a matching delimiter is found, the intervening span is wrapped into a temporary intermediate constituent whose label is marked with an \texttt{@} prefix.
If a delimiter is unpaired, it is attached to the closest structural neighbor using a deterministic heuristic:
left delimiters are attached to the following sibling, and right delimiters are attached to the preceding sibling.
This normalization step ensures that punctuation is structurally subordinated before binarization and reduces variation arising from punctuation attachment decisions.

\paragraph{Head-driven binarization}
After punctuation-aware normalization, we apply a deterministic head-driven binarization procedure to convert $n$-ary branching to binary branching.
For each non-terminal node, the procedure identifies the designated head child and builds a binary structure by composing non-head siblings around it.
For nodes with branching factor $k>2$, the algorithm iteratively groups siblings into intermediate constituents (labeled with an \texttt{@} prefix) while maintaining the head child as the final attachment point within the local binarized configuration.
A post-processing step collapses unary non-terminal chains introduced by binarization.
The resulting binarization is designed to be reversible under deterministic debinarization, subject to the marking scheme of intermediate nodes.

\paragraph{Parser training}
We train neural constituency parsers using the Stanza framework on the binarized trees.
For English, we use a RoBERTa-base encoder, and for Chinese, we use \texttt{bert-base-chinese}.
Encoders are fine-tuned during training (\texttt{--bert\_finetune}).
To preserve consistency with the preprocessing pipeline, automatic retagging is disabled (\texttt{--no\_retag}), so the parser uses the gold part-of-speech tags provided by the treebanks.
Training is run for 100 epochs.
Treebank utilities are integrated via Stanford CoreNLP 4.5.10.

\section{Experiment 3 details} \label{experiment-3-detail}

This section specifies the deterministic conversion procedures used in Experiment~3.
The only learned component is the head chooser \(H\), which selects a head child index for each non-terminal node.
All remaining steps are fixed and shared across the rule-based and learned-headedness conditions.

\paragraph{Constituency to dependency}

Given an ordered constituency tree \(C\) whose preterminals correspond to the token sequence \(w_1,\ldots,w_n\), we assign token indices \(1..n\) by a left-to-right traversal.
The conversion constructs a dependency structure represented by arrays \(\textsc{head}[1..n]\) and \(\textsc{rel}[1..n]\).
The procedure is defined by a single recursive pass over \(C\) (Algorithm~\ref{alg:const2dep_abstract}).
For each subtree \(v\), the recursion returns the head token id \(h(v)\) of that subtree.
If \(v\) is preterminal, \(h(v)\) is its token id.
Otherwise, let \(\mathit{ch}(v)=\langle c_1,\ldots,c_k\rangle\) be the ordered children and let \(p=H(v)\) be the predicted head-child position.
The head token of the subtree is then \(h(v)=h(c_p)\).
For each non-head child \(c_j\) with \(j\neq p\), we attach the head token of that child subtree to \(h(v)\) by setting
\(\textsc{head}[h(c_j)]\gets h(v)\).
After the recursion returns at the root, we assign the root head token \(r=h(\mathrm{root}(C))\) to the artificial root by setting \(\textsc{head}[r]\gets 0\) and \(\textsc{rel}[r]\gets\texttt{root}\).

\begin{algorithm}[!ht]
\caption{\textsc{Constituency-to-dependency}}
\label{alg:const2dep_abstract}
\begin{algorithmic}[1]
\STATE \textbf{Input:} ordered constituency tree \(C\); head chooser \(H(v)\in\{1,\ldots,|\mathit{ch}(v)|\}\)
\STATE \textbf{Output:} dependency head array \(\textsc{head}[1..n]\) and relation array \(\textsc{rel}[1..n]\)

\STATE Assign token indices \(1..n\) to the preterminals of \(C\) in left-to-right order
\STATE Initialize \(\textsc{head}[i]\gets\bot\), \(\textsc{rel}[i]\gets\bot\) for all \(i\)

\STATE \textbf{function} Collapse(\(v\)) \COMMENT{returns head token id of subtree \(v\)}
\begin{ALC@g}
  \STATE \textbf{if} \(v\) is a preterminal with token id \(t\) \textbf{then return} \(t\) \textbf{end if}
  \STATE For each child \(c\in\mathit{ch}(v)\): \(h_c \gets\) Collapse(\(c\))
  \STATE \(p \gets H(v)\); \(h \gets h_{c_p}\) \COMMENT{head token of head child}
  \STATE \textbf{for each} child \(c_j \neq c_p\) \textbf{do}
  \begin{ALC@g}
    \STATE \textsc{head}\([h_{c_j}]\gets h\)
  \end{ALC@g}
  \STATE \textbf{end for}
  \STATE \textbf{return} \(h\)
\end{ALC@g}
\STATE \textbf{end function}

\STATE \(r \gets\) Collapse(root(\(C\)))
\STATE \textsc{head}\([r]\gets 0\); \textsc{rel}\([r]\gets\texttt{root}\)
\STATE \textbf{return} \((\textsc{head},\textsc{rel})\)
\end{algorithmic}
\end{algorithm}


\paragraph{Preprocessing and reversibility assumptions}

The conversion may be applied either to original trees or to trees that contain binarization marker nodes (e.g., labels prefixed with \texttt{@}).
When marker nodes are present, we optionally debinarize by splicing their children into the parent prior to conversion, ensuring that head selection is evaluated on comparable n-ary configurations.
All label normalizations used by \(H\) and by \(\textsc{ComposeRel}\) are deterministic and fixed for all experiments.

\paragraph{Evaluation protocol}

We evaluate the derived dependency structures intrinsically against gold dependencies using unlabeled attachment score (UAS), so that results reflect head attachment decisions independently of relation label choices.
The same conversion procedure is run under two head sources: (i) rule-based head percolation and (ii) learned headedness.
Because the conversion is otherwise deterministic, differences in UAS are attributable to differences in constituent head assignment.
Constituency reconstruction from the derived dependencies is also deterministic; reconstruction quality is evaluated using labeled span \(F_1\) against the original gold constituency trees.

\end{document}